\documentclass[twocolumn,9pt]{IEEEtran}

\usepackage{amsmath,graphicx}
\usepackage{cite,amssymb,amsfonts,color}

\usepackage{bm}
\usepackage{algorithmic}
\usepackage{algorithm}
\usepackage{multicol}
\makeatletter
\newcommand*{\rom}[1]{\expandafter\@slowromancap\romannumeral #1@}
\makeatother

\begin{document}
\title{Segment Parameter Labelling in MCMC Mean-Shift  Change Detection }

\author{Alireza Ahrabian and Shirin Enshaeifar and Clive Cheong-Took  and Payam Barnaghi\\
\thanks{The authors Alireza Ahrabian, Shirin Enshaeifar and Payam Barnaghi are with the Department of Electrical and Electronic Engineering, Institute for
Communication Systems and Clive Cheong-Took is with the Department of Computer Science at the University of Surrey, GU2 7HX, U.K. (email: \{a.ahrabian,  s.enshaeifar, c.cheongtook and  p.barnaghi\}@surrey.ac.uk.}} 
\maketitle 
\begin{abstract}
This work addresses the problem of segmentation in  time series data with respect to a statistical parameter of interest in Bayesian models. It is common to assume that the  parameters are distinct  within each segment. As such, many Bayesian change point detection models do not exploit the segment parameter patterns, which can improve performance. This work proposes a Bayesian mean-shift change point detection algorithm that makes use of repetition in  segment parameters, by introducing segment class labels that utilise a Dirichlet process prior. The performance of the proposed approach was assessed on both synthetic and real world data, highlighting  the enhanced performance when using parameter labelling. 
\end{abstract}
\section{Introduction}
The partitioning of time series data into segments of piecewise stationary statistics is important in many fields ranging from the analysis of accelerometer data corresponding to human gait motion to the analysis of  biomedical data \cite{Sejdic09}. Bayesian approaches that consider the change point transition times and number of segments as statistical parameters are particularly  interesting. Such techniques  enable  the modeller to incorporate uncertainty in the statistical parameters being estimated, thereby providing  the means of controlling  the model complexity with respect to the model fit. 
\\
\indent The work in \cite{Punskaya02} proposed  a fully hierarchical  Bayesian model where  the uncertainty in the relevant parameters of interest were captured using appropriate prior probabilities. A reversible jump Markov Chain Monte Carlo (MCMC) technique was then used in order to obtain estimates of the relevant parameters. While  closed form analytical expressions to the posterior derived in \cite{Punskaya02} is not possible, the work in \cite{Fearnhead05} proposed an efficient recursive  solution in order to compute such an estimate. More recently, a multivariate extension that captures  changes in the dependency structure of data was proposed in \cite{Xuan07}. While the work in \cite{Hensley17} proposed a nonparametric Bayesian method for detecting changes in the variance of data. 
\\
\indent  Many change point detection algorithms often assume that parameters from different segments are distinct, however,  infinite hidden Markov models (IHMM)  and Dirichlet process mixture models (DPMM) assume that data at each time point can be generated by a parameter that belongs to a potentially infinite number of  states or classes that is determined from the data set \cite{Rasmussen00}\cite{Fox11}. In particular  these methods have been introduced  into change point detection algorithms \cite{Ko15}. However, such work often assigns a parameter (belonging to a particular state) label to each time point and not the parameters corresponding to a given segment. 
\\
\indent  In this paper we propose a mean-shift (that can be generalised to parameters of a statistical   model) Bayesian change point  detection algorithm that exploits repetition in segment parameters for more robust segmentation. This is achieved by extending the change point detection algorithm introduced in \cite{Punskaya02}, by including parameter class labels that employ a Dirichlet process prior for identifying  the number of distinct segment parameters. Simulations on synthetic  and real world data verify the efficacy  of the proposed method.
\section{Background}
\subsection{MCMC Change Point Detection}
  Given a set of $N$ data points $\mathbf{x}$  and transition times $\boldsymbol{\tau}_{K}=[\tau_{1},...,\tau_{K}]$ where $\tau_{0}=1$ and $\tau_{K+1}=N$, there exist $K+1$ segments such that for each segment the following functional relationship  between the data points (within the time indices $\tau_{i}+1\leq \tau \leq \tau_{i+1}$) and  the statistical parameter $\phi_{i}$ is satisfied, that is 
\begin{equation}
\boldsymbol{\text{x}}_{\tau_{i}+1:\tau_{i+1}}=f(\boldsymbol{\text{x}}_{\tau_{i}+1:\tau_{i+1}},\phi_{i}) + \boldsymbol{\text{n}}_{\tau_{i}+1:\tau_{i+1}}
\label{model}
\end{equation}
for $i=\{0,\dots,K\}$, where $\boldsymbol{\text{n}}_{\tau_{i}+1:\tau_{i+1}}$ is a set of i.i.d.  zero mean Gaussian noise samples (with a  specific variance of $\sigma_{i}^{2}$). The hierarchical Bayesian model introduced in \cite{Punskaya02}, derived a posterior distribution such that the target parameters of interest were both $K$ and $\boldsymbol{\tau}_{K}$, namely 
\begin{equation}
p(K,\boldsymbol{\tau}_{K},\boldsymbol{\phi},\mathcal{H}|\mathbf{x})\propto p(\mathbf{x}|\boldsymbol{\phi},K,\boldsymbol{\tau}_{K})p(\boldsymbol{\phi},K,\boldsymbol{\tau}_{K}|\mathcal{H})p(\mathcal{H})
\label{post_mcmc}
\end{equation}
 where $\boldsymbol{\phi}$ corresponds to the vector of segment parameters (that is integrated out of the posterior) and $\mathcal{H}$ represents the set of  hyperparameters for the relevant  prior probabilities. The likelihood function used in the posterior distribution  \eqref{post_mcmc}  assumes that the parameters $\phi_{i}$ are distinct for each segment \cite{Punskaya02}, that is  
\begin{equation}
p(\mathbf{x}|\boldsymbol{\phi},K,\boldsymbol{\tau}_{K})=\prod_{i=0}^{K}p(\mathbf{x}_{\tau_{i}+1:\tau_{i+1}}|\phi_{i})
\label{like_mcmc}
\end{equation}
 \subsection{Dirichlet Process Mixture Model}
 A finite mixture model (FMM) assumes that the data is  drawn from a weighted combination of  distributions from the same parametric family with differing parameters \cite{Rasmussen00}\cite{Neal00}. That is, $p(\mathbf{x})=\sum_{v=1}^{V}\pi_{v}f_{p}(\mathbf{x}|\theta_{v})$,  where $V$ is the number of classes, $\pi_{v}$ is the mixing coefficient and $\theta_{v}$ corresponds to the class parameter/s of the probability distribution $f_{p}(.)$. For a given data set, determining the number of classes $V$ in a systematic way can be a challenging task. In order to overcome this problem, we must first consider the generative model of the FMM, where the $i^{\text{th}}$ class indicator random variable $c_{i}$ is introduced, that is 

\begin{equation}
\begin{aligned}
x_{i}|c_{i},\boldsymbol{\theta}&\sim f(x_{i}|\theta_{c_{i}})\\
c_{i}|\boldsymbol{\pi}&\sim \text{Discrete}(\pi_{1},\dots,\pi_{V})\\
\theta_{v}&\sim G_{0}\\
 \boldsymbol{\pi} &\sim \text{Dir}(\alpha/V,\dots,\alpha/V)
\end{aligned}
\label{gen_model_dirichlet}
\end{equation}
where $G_{0}$ corresponds to the 
 prior distribution  of the parameters, $\boldsymbol{\theta}=[\theta_{1},\dots,\theta_{V}]$ and $\boldsymbol{\pi}=[\pi_{1},\dots,\pi_{V}]$. The mixing coefficients $\boldsymbol{\pi}$ in the model  \eqref{gen_model_dirichlet} govern the likelihood of selecting a given class. By employing a Dirichlet distribution prior on the mixing coefficients and taking the limit $V\rightarrow \infty$, results in the Dirichlet process (DP) mixture model \cite{Neal00}, which is often written as 
\begin{equation}
\begin{aligned}
x_{i}|\theta_{i}&\sim f(x_{i}|\theta_{i})\\
\theta_{i}&\sim G\\
G&\sim \text{DP}(G_{0},\alpha)
\end{aligned}
\label{DP}
\end{equation}
where $G$ is drawn from the Dirichlet process with base measure $G_{0}$.
Inference of the DP mixture model is generally carried out using Gibbs sampling, where the state of the Markov chain consists of all the parameters and class labels. A particularly interesting outcome of the DP mixture model is in the inference of the class parameters (which does not require the direct  specification  of the number of classes). That is, the conditional posterior probability of assigning a data point to  an existing class is given by 
\vspace{-0mm}
\begin{equation}
p(c_{i}=v|\boldsymbol{\text{c}}_{-i},x_{i},\boldsymbol{\theta})\propto\frac{n_{-i,v}}{N-1+\alpha}L(x_{i}|\theta_{v})
\label{cond_post_dirichlet1}
\end{equation}
where $n_{-i,v}$ is the number of data points (excluding $x_{i}$) assigned to class $v$ and $\boldsymbol{\text{c}}_{-i}$ is a vector of class labels excluding  $c_{i}$. While the conditional class posterior probability for assigning the data point $x_{i}$ to a new class is given by
\begin{equation}
\begin{aligned}
p(c_{i}\neq c_{l} \quad \hspace{-2mm}&\text{for all}\quad \hspace{-2mm} i\neq l|\boldsymbol{\text{c}}_{-i},x_{i})\\
&\propto\frac{\alpha}{N-1+\alpha}\int L(x_{i}|\boldsymbol{\theta})dG_{0}(\boldsymbol{\theta})
\label{cond_post_dirichlet2}
\end{aligned}
\end{equation}

\begin{figure}[t!]
  \centering
\includegraphics[width=1\columnwidth]{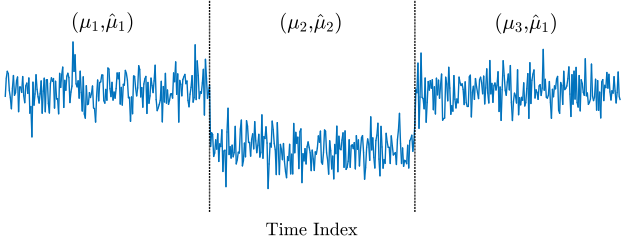}
        \caption{Figure illustrating the difference between  distinct mean $\mu_{i}$ and class mean $\hat{\mu}_{i}$ within each segment. }\label{fig:Mean}
\end{figure}

\section{Proposed Work}
We propose a novel mean-shift change detection algorithm  by including parameter class labels in the technique proposed in \cite{Punskaya02}. That is, given a set of data points $\boldsymbol{\text{x}}$ and transition  times $\boldsymbol{\tau}_{K}$, where the functional relationship defined in \eqref{model} for mean-shift change detection (assuming distinct parameters in each segment \cite{Punskaya02}) is given by  $f(\boldsymbol{\text{x}}_{\tau_{i}+1:\tau_{i+1}},\mu_{i}) =\mu_{i}\boldsymbol{1}_{1:\tau_{i+1}-\tau_{i}}$, with $\mu_{i}$ corresponding to the mean of the segment and $\boldsymbol{1}_{1:\tau_{i+1}-\tau_{i}}$ is a vector with elements equal to 1. Each segment has  a distinct parameter $\mu_{i}$;  however, data often exhibit parameters that repeat across different segments, therefore the model in \cite{Punskaya02} does not efficiently use the similarity  in the segment parameters.
\\
 \indent We propose  to include parameter class label $c_{i}$, such that the mean parameters $\boldsymbol{\mu}=[\mu_{0},\dots,\mu_{K}]$, are generated by the following Gaussian mixture model
\begin{equation}
p(\boldsymbol{\mu}|\hat{\boldsymbol{\mu}},\hat{\boldsymbol{\sigma}}^{2},\boldsymbol{\pi})=\sum_{v=1}^{V}\pi_{v}\mathcal{N}(\boldsymbol{\mu}|\hat{\mu}_{v},\hat{\sigma}^{2}_{v})
\label{proposed_mix}
\end{equation}
\noindent where  $\hat{\boldsymbol{\mu}}=[\hat{\mu}_{1},\dots,\hat{\mu}_{V}]$ and $\hat{\boldsymbol{\sigma}}^{2}=[\hat{\sigma}^{2}_{1},\dots,\hat{\sigma}^{2}_{V}]$ correspond to the class parameters. For each segment the functional relationship in \eqref{model}  is given by $f(\boldsymbol{\text{x}}_{\tau_{i}+1:\tau_{i+1}},\hat{\mu}_{i}) =\hat{\mu}_{c_{i}}\boldsymbol{1}_{1:\tau_{i+1}-\tau_{i}}$, where different segments now can be assigned to the same class of parameters (see Fig.~\ref{fig:Mean}) . As a result, data in segments with the same parameter labels are combined for more robust segmentation. 
\subsection{Bayesian Model}
The following model formally states  the proposed mean-shift change point algorithm that includes parameter labelling
\begin{equation}
\begin{aligned}
\boldsymbol{\text{x}}_{\tau_{i}+1:\tau_{i+1}}|\hat{\mu}_{i},\sigma^{2}_{i}&\sim f_{j}(\boldsymbol{\text{x}}_{\tau_{i}+1:\tau_{i+1}}|\hat{\mu}_{i},\sigma^{2}_{i})\\
\sigma^{2}_{i}&\sim G_{\sigma^{2}}\\
\mu_{i}|\hat{\mu}_{i},\hat{\sigma}^{2}_{i}&\sim \mathcal{N}(\mu_{i}|\hat{\mu}_{i},\hat{\sigma}^{2}_{i})\\
(\hat{\mu}_{i},\hat{\sigma}^{2}_{i})&\sim G\\
G&\sim \text{DP}(G_{0},\alpha)\\
\boldsymbol{\text{x}}_{\tau_{i}+1:\tau_{i+1}}|\boldsymbol{\tau}_{K},\mu_{i}&\sim f_{j}(\boldsymbol{\text{x}}_{\tau_{i}+1:\tau_{i+1}}|\mu_{i})\\
 \boldsymbol{\tau}_{K}, K &\sim \text{Bin}(\boldsymbol{\tau}_{K}, K |\lambda)
\end{aligned}
\label{gen_model_changepoint}
\end{equation}
where $\text{Bin}(.)$ corresponds to a Binomial distribution, $G_{0}$ is the joint  prior distribution of the class mean and variance, $G_{\sigma^{2}}$ is the prior distribution of the variance of the data points with the same class label and $f_{j}(.)$ corresponds to the joint Normal distribution. 
\\
\indent The state of the Markov chain consists  of the following parameters, $\{\boldsymbol{\tau}_{K},K,\boldsymbol{\text{c}}_{K},\hat{\boldsymbol{\mu}},\hat{\boldsymbol{\sigma}}^{2},\boldsymbol{\sigma}^{2}\}$, where $\boldsymbol{\text{c}}_{K}=[c_{0},\dots,c_{K}]$ and $\boldsymbol{\sigma}^{2}=[\sigma_{1}^{2},\dots,\sigma_{V}^{2}]$. Inference of the parameters is carried out by using  a  Metropolis-Hastings-within-Gibbs sampling scheme. The Gibbs moves are performed on each parameter in the set, $\{\boldsymbol{\text{c}}_{K},\hat{\boldsymbol{\mu}},\hat{\boldsymbol{\sigma}}^{2},\boldsymbol{\sigma}^{2}\}$, while a variation of the Metropolis-Hastings algorithm is used to obtain samples for the parameters $\{\boldsymbol{\tau}_{K},K\}$.
\\
\indent The Gibbs sampling procedure requires the conditional posterior  distributions for all class means $\hat{\mu}_{v}$ and variances $\hat{\sigma}^{2}_{v}$, along with  the conditional  posterior variance of the data points within the same class $\sigma^{2}_{v}$ and each class label $c_{i}$. The exact derivation of these conditional  posterior distributions  were carried out in both \cite{Punskaya02} and \cite{Rasmussen00}; where  we have assumed the following class mean prior, $p(\hat{\mu}_{v}|\lambda,\delta)\sim \mathcal{N}(\lambda,\delta\sigma^{2}_{v})$,  that is dependent on $\sigma^{2}_{v}$. Furthermore, the class variance has an inverse Gamma prior given by, $p(\hat{\sigma}^{2}_{v}|\beta,\omega)\sim \mathcal{IG}(\beta,\omega)$. 
\\
\indent The conditional  posterior distribution of the parameters $\{\boldsymbol{\tau}_{K},K\}$ is given by, $p(\boldsymbol{\tau}_{K},K|\lambda,\boldsymbol{\text{c}}_{K},\hat{\boldsymbol{\mu}},\boldsymbol{\sigma}^{2},\boldsymbol{\text{x}})$. Owing to the selection of the appropriate conjugate priors, we can integrate out the nuisance parameters $\{\hat{\boldsymbol{\mu}},\boldsymbol{\sigma}^{2},\lambda\}$. This is carried out by considering the following posterior distribution
\begin{equation}
\begin{aligned}
&p(\boldsymbol{\hat{\mu}},\boldsymbol{\sigma}^{2},\boldsymbol{\tau}_{K},K,\lambda|\boldsymbol{\text{c}}_{K},\mathbf{x})\propto p(\mathbf{x}|\boldsymbol{\hat{\mu}},\boldsymbol{\sigma}^{2},K,\boldsymbol{\tau},\boldsymbol{\text{c}}_{K})\\
&\times p(K,\boldsymbol{\tau}|\lambda)p(\lambda)\prod_{v=1}^{V}p(\hat{\mu}_{v}|\lambda,\delta)p(\sigma^{2}_{v}|\nu,\gamma)
\label{posterior_proposed}
\end{aligned}
\end{equation}
where $p(\boldsymbol{\tau}_{K},K|\lambda)=\lambda^{K}(1-\lambda)^{T-K-1}$, $p(\sigma^{2}_{v}|\nu,\gamma)\sim \mathcal{IG}(\nu,\gamma)$ and $p(\lambda)$ has uniform probability between $[0,1]$.  The likelihood function is given by 
\begin{equation*}
p(\mathbf{x}|\boldsymbol{\hat{\mu}},\boldsymbol{\sigma}^{2},K,\boldsymbol{\tau}_{K},\boldsymbol{\text{c}}_{K})=\prod_{v=1}^{V}\prod_{i:c_{i}=v}p(\mathbf{x}_{\tau_{i}+1:\tau_{i+1}}|\hat{\mu}_{v},\sigma_{v}^{2})
\label{likelihood_proposed}
\end{equation*}
where data within segments with the same parameter label $v$ are combined for potentially more accurate parameter estimation (in the mean squared error sense). Integration of \eqref{posterior_proposed} with respect to the parameters $\{\hat{\mu}_{v},\sigma_{v}^{2},\lambda\}$ results in the following expression for the conditional posterior distribution of the parameters $\{\boldsymbol{\tau}_{K},K\}$
\begin{equation}
\begin{aligned}
&p(\boldsymbol{\tau}_{K},K|\boldsymbol{\text{c}}_{K},\mathbf{x})\propto \prod_{v=1}^{V}\frac{2^{\frac{\nu}{2}}}{\Gamma(\frac{\nu}{2})}\Gamma(K+1)\Gamma(N-K+1) \left(\frac{\gamma}{2}\right)^{\frac{\nu}{2}} \\
& \times\Gamma\left(\frac{d_{v}+\nu}{2}\right)\pi^{-\frac{d_{v}}{2}}\left[\gamma+Y^{T}_{v}P_{v}Y_{v}\right]^{-\frac{d_{v}+\nu}{2}}(d_{v}+\delta^{-1})^{-\frac{1}{2}}
\end{aligned}
\label{full_posterior_proposed}
\end{equation}
where $Y_{v}$ is the concatenated vector of all data points with the same segment  label $v$, $d_{v}$ is the number of data points with label $v$, and  $P_{v}=\left(\mathbf{I}_{d_{v}}-\boldsymbol{1}_{1:d_{v}}M_{v}\boldsymbol{1}_{1:d_{v}}^{T}\right)$, with $M_{v}=(d_{v}+\delta^{-1})^{-\frac{1}{2}}$. Finally, we note that there are some challenges from drawing samples from \eqref{full_posterior_proposed} due to the dependence on $\boldsymbol{\text{c}}_{K}$ that we have addressed in the next section.
\subsection{MCMC Sampling}

Samples for  parameters, $\{\boldsymbol{\text{c}}_{K},\hat{\boldsymbol{\mu}},\hat{\boldsymbol{\sigma}}^{2},\boldsymbol{\sigma}^{2}\}$ are obtained by drawing samples from the following posterior densities: $p(\hat{\mu}_{v}|\boldsymbol{\mu},\boldsymbol{\text{c}}_{K},\hat{\sigma}^{2}_{v},\sigma^{2}_{v})$, $p(\hat{\sigma}^{2}_{v}|\boldsymbol{\mu},\boldsymbol{\text{c}}_{K},\hat{\mu}_{v})$, $p(c_{i}|\boldsymbol{\text{c}}_{-i},\boldsymbol{\mu},\hat{\mu}_{v},\hat{\sigma}^{2}_{v})$ and $p(c_{i}\neq c_{l} \quad \hspace{-2mm}\text{for all}\quad \hspace{-2mm} i\neq l|\boldsymbol{\text{c}}_{-i},\boldsymbol{\mu})$ where details on exact distribution form and sampling  are found in \cite{Rasmussen00}.  While $p(\sigma^{2}_{v}|\mathbf{x},\boldsymbol{\tau}_{K},\boldsymbol{\text{c}}_{K})$ is found by concatenating all the data points that have the same segment label $v$. More details on exact distribution form and sampling  can be found in \cite{Punskaya02}. 
\\
\indent In order to evaluate the conditional posterior distribution $p(\boldsymbol{\tau}_{K},K|\boldsymbol{\text{c}}_{K},\mathbf{x})$ we use a modification of the  Metropolis-Hastings algorithm   outlined in \cite{Punskaya02} that incorporates  segment labels $\boldsymbol{\text{c}}_{K}$. Given the current state of the Markov chain $\{\boldsymbol{\tau}_{K},K\}$,  we select  one  of the steps with the following probabilities:

\begin{itemize}
  \item  birth of a change point with probability, $b$

  \item  death of a change point with probability, $d$

 \item update of change point positions with probability, $u$
\end{itemize}

where   $b=d=u$ for $0<K<K_{max}$, and $b+d+u=1$ for $0\leq K\leq K_{max}$. 
\\
\indent A birth  move consists of proposing a new change point  $\tau_{prop}$ with uniform  probability from the existing time indices $[2,N-1]$ excluding the time indices $\boldsymbol{\tau}_{K}$. The proposed set of  change points including $\tau_{prop}$ is given by $\boldsymbol{\tau}_{K+1}$, where the segment between the time indices $[\tau_{i},\tau_{i+1}]$ ($\tau_{i}<\tau_{prop}<\tau_{i+1}$) with the class variable $c_{i}$ is split into two new segments with  two new class variables  $\{\hat{c_{i}},\hat{c}_{i+1}\}$. As we have not yet inferred the new class labels from the conditional  class posterior distributions,  we assume that the two classes $\{\hat{c_{i}},\hat{c}_{i+1}\}$ are distinct and thus independent  from all other segments, to circumvent  the lack of information we have for assignment to an existing class.  The proposed transition  time is accepted with the following  probability,  $\alpha_{birth}=\text{min}\{1,r_{birth}\}$, where
{\small{
\begin{equation*}
r_{birth}=\frac{p(\boldsymbol{\tau}_{K+1},K+1|\boldsymbol{\text{c}}_{K+1},\mathbf{x})}{p(\boldsymbol{\tau}_{K},K|\boldsymbol{\text{c}}_{K},\mathbf{x})}\frac{q(\boldsymbol{\tau}_{K}|\boldsymbol{\tau}_{K+1})q(K|K+1)}{q(\boldsymbol{\tau}_{K+1}|\boldsymbol{\tau}_{K})q(K+1|K)}
\label{accept_probab}
\end{equation*}
}}
with  $q(\boldsymbol{\tau}_{K+1}|\boldsymbol{\tau}_{K})=\frac{1}{N-K-2}$, $q(K+1|K)=b$, $q(\boldsymbol{\tau}_{K}|\boldsymbol{\tau}_{K+1})=\frac{1}{K+1}$ and $q(K|K+1)=d$.
\\
\indent The death move proposes to remove a transition time $\tau_{prop}$, by choosing with uniform probability from the set $[\tau_{1},\dots,\tau_{K}]$. That is, the segments $\tau_{i}+1\leq \tau\leq \tau_{i+1}$ and $\tau_{i+1}+1\leq \tau\leq \tau_{i+2}$ where $\tau_{i+1}=\tau_{prop}$, are combined into one segment $\tau_{i}+1\leq \tau\leq \tau_{i+2}$. Furthermore, the class labels $\{c_{i},c_{i+1}\}$ are combined into one segment with a new class label (utilising the argument used for the birth  of a change point), that is $\hat{c}_{i}\neq c_{j}$ for all $j\neq i$. The removal of $\tau_{prop}$ is accepted with probability   $\alpha_{death}=\text{min}\{1,r_{birth}^{-1}\}$.
\\
\indent The update of the change points is carried by first removing the time index  $\tau_{j}$ in $\boldsymbol{\tau}_{K}$ and proposing a new change point at some new location. That is, the death move is first applied followed by  a birth move, for all $ j=\{1,\dots,K\}$.

\section{Simulations}

\subsection{Synthetic Data}

We evaluated the performance  of the proposed algorithm under two different scenarios. Namely, the first scenario considered that every segment was produced with a randomly generated mean parameter (with fixed variance), while the second scenario assumed that each segment has a mean parameter (with fixed variance) drawn from a fixed number of repeating classes. Furthermore, each segment length was drawn with uniform probability  between $[20,70]$ samples, while the number of segments were also selected with a random probability (each realisation on average had approximately 7 change points). The proposed method was compared with the following algorithms: MCMC \cite{Punskaya02}, Group Fused LASSO \cite{Bleakley11} and PELTS \cite{Killick12}. The following parameters were selected for the proposed method: $\alpha=2$, $\lambda=0$, $\delta=1$, $\nu=2$, $\gamma=20$, $\beta=0.01$ and $\omega=200$ (in general we fix all the parameters except $\alpha$ and $\gamma$); we note that parameters for each method were selected  such that the proportion of false positives were as close as possible for each method. We evaluated the performance of the respective algorithms using  the  following  measures:  the  proportion  of    true  positives  along  with  the  absolute  error  in the change point location estimate. 
\\
\indent From Table 1 it can be observed that the proposed method outperformed both the MCMC in \cite{Punskaya02} along with the Group Fused LASSO, with respect to the number of true positives and change point location estimation, for time series data with random mean parameter assignment. Furthermore,  Table 2 shows that the proposed method was able to significantly outperform the MCMC and Group Fused LASSO algorithms with respect to the proportion of true positives when segmenting data with repeating mean parameters. However, the PELTS algorithm was able to outperform the proposed method with respect to the proportion  of true positives and change point location estimation (as shown in both Table 1 and 2). The PELTS algorithm represents the state of the art for detecting changes with respect to a fixed statistical parameter (using dynamic programming), whereas the proposed Bayesian change point detection algorithm can cater for linear statistical models (e.g. autoregressive  model) with arbitrary  model orders for each segment, along with the prediction of change points using the predictive posterior distribution  which is not possible with PELTS. 

\begin{table}[h!]
\caption{Random mean parameter assignment.}

\begin{center}
\small\addtolength{\tabcolsep}{-5pt} { \small
\begin{tabular}{|c| c c c| }   \hline 
 ~~~Methods ~~~ & ~True Positives ~ & ~False Positives~  & ~ Error ~ 
\\ \hline
~~Proposed Method~~ &$97.3\%$ &$5.9\%$   &$1.9$  \\
               \hline
MCMC \cite{Punskaya02}~~~~   &$84.6\%$ &$9.6\%$  & $3.16$ \\
                 \hline
G. F. LASSO \cite{Bleakley11} &$63.4\%$ &$50.6\%$& $5.52$  \\                        
 \hline
PELTS \cite{Killick12}~~~~  &$99.9\%$ &$6.2\%$& $1.4$  \\                        
 \hline
\end{tabular}}
\end{center}
\label{Tab:results}
\end{table}

\begin{table}[h!]
\caption{Repeating mean parameter assignment.}

\begin{center}
\small\addtolength{\tabcolsep}{-5pt} { \small
\begin{tabular}{|c| c c c| }   \hline 
 ~~~Methods ~~~ & ~True Positives ~ & ~False Positives~  & ~ Error ~ 
\\ \hline
~~Proposed Method~~ &$85.9\%$ &$5.7\%$   &$3.16$  \\
               \hline
MCMC \cite{Punskaya02}~~~~   &$36.3\%$ &$7.5\%$  & $5.38$ \\
                 \hline
G. F. LASSO \cite{Bleakley11} &$41.6\%$ &$31.6\%$& $4.6$  \\                        
 \hline
PELTS \cite{Killick12}~~~~  &$93.2\%$ &$5.9\%$& $3.35$  \\                        
 \hline
\end{tabular}}
\end{center}
\label{Tab:results}
\end{table}

\section{Conclusions and Future Work}
This work proposes a mean-shift change point detection algorithm that captures parameter repetition for more robust time series segmentation. This was achieved by labelling parameters for a given segment and by employing a Dirichlet process prior. We have shown the advantage of the proposed method on synthetic and real world data. Future work will extend the proposed method for a wider class of time series models as well as including hyperparameter updates. 

\bibliographystyle{IEEEtran}
\bibliography{journals}

\end{document}